\documentclass[sigconf, screen, authorversion, nonacm]{acmart}

\definecolor{minor}{HTML}{FF0000}
\usepackage{xcolor}

\usepackage{graphicx}
\usepackage{csquotes}
\usepackage{makecell}
\usepackage{subcaption}
\usepackage{svg}
\usepackage{enumitem}
\usepackage{hyperref}

\usepackage[many]{tcolorbox}    	
\usepackage{setspace}               
\usepackage{multicol}               

\makeatletter
\newcommand{\setword}[2]{%
  \phantomsection
  #1\def\@currentlabel{\unexpanded{#1}}\label{#2}%
}
\makeatother

\AtBeginDocument{%
  \providecommand\BibTeX{{%
    \normalfont B\kern-0.5em{\scshape i\kern-0.25em b}\kern-0.8em\TeX}}}
 
\setcopyright{acmcopyright}
\copyrightyear{2018}
\acmYear{2018}
\acmDOI{10.1145/1122445.1122456}

\acmConference[Woodstock '18]{Woodstock '18: ACM Symposium on Neural
  Gaze Detection}{June 03--05, 2018}{Woodstock, NY}
\acmBooktitle{Woodstock '18: ACM Symposium on Neural Gaze Detection,
  June 03--05, 2018, Woodstock, NY}
\acmPrice{15.00}
\acmISBN{978-1-4503-XXXX-X/18/06}

\newtcolorbox{boxA}{
    boxrule = 1pt,
    colframe = black 
}

\begin{document}

\title{Prompting a Large Language Model to Generate Diverse Motivational Messages}
\subtitle{A Comparison with Human-Written Messages}


\author{Samuel Rhys Cox}
\email{samcox@comp.nus.edu.sg}
\orcid{0000-0002-4558-6610}
\affiliation{%
  \institution{National University of Singapore}
    \country{} 
  }

\author{Ashraf Abdul}
\email{ashrafabdul@pm.me}
\orcid{0000-0002-3383-6440}
\affiliation{\institution{Unaffiliated}
  \country{} 
  }

\author{Wei Tsang Ooi}
\email{ooiwt@comp.nus.edu.sg}
\orcid{0000-0001-8994-1736}
\affiliation{\institution{National University of Singapore}
  \country{} 
  }

\begin{abstract}

Large language models (LLMs) are increasingly capable and prevalent, and can be used to produce creative content. 
The quality of content is influenced by the prompt used, with more specific prompts that incorporate examples generally producing better results.
On from this, it could be seen that using instructions written for crowdsourcing tasks (that are specific and include examples to guide workers) could prove effective LLM prompts.
To explore this, we used a previous crowdsourcing pipeline that gave examples to people to help them generate a collectively diverse corpus of motivational messages.
We then used this same pipeline to generate messages using GPT-4, and compared the collective diversity of messages from: (1) crowd-writers, (2) GPT-4 using the pipeline, and (3 \& 4) two baseline GPT-4 prompts.
We found that the LLM prompts using the crowdsourcing pipeline caused GPT-4 to produce more diverse messages than the two baseline prompts.
We also discuss implications from messages generated by both human writers and LLMs.

\end{abstract}



\maketitle


\section{Introduction}

\begin{figure}
    \begin{boxA}
    {\footnotesize \textbf{Simple-GPT prompt:}\\
    You will now write 250 messages to motivate someone to exercise. Please use the following guidelines:\\
    - Write concise messages (imagine the message could be sent to people on a smartphone). Messages should generally be 1 to 3 sentences long.\\
    - Output your response with one message per line.
    \par\noindent\rule{\textwidth}{0.5pt}
    \textbf{Diverse-Naïve-GPT prompt:}\\
    You will now write 250 messages to motivate someone to exercise. Please use the following guidelines:\\
    - Write concise messages (imagine the message could be sent to people on a smartphone). Messages should generally be 1 to 3 sentences long.\\
    - Ensure the messages are diverse from one another, by limiting message repetition and similarity (such as by limiting repetition of themes or phrasing).\\
    - Output your response with one message per line.
    \par\noindent\rule{\textwidth}{0.5pt}
    \begin{spacing}{1}
    \textbf{Phrase-GPT prompt:}\\
    You will now write 250 messages to motivate someone to exercise, and you will be shown short phrases to inspire you. Please use the following guidelines:\\
    - Use the phrases for inspiration to write a message that would motivate someone to exercise.\\
    - Write concise messages (imagine the message could be sent to people on a smartphone). Messages should generally be 1 to 3 sentences long.\\
    - You can write using the entire phrases, or fragments of phrases.\\
    - Ignore phrases that you don't find relevant to writing motivational messages for exercise (but feel free to be creative if you can use any of the phrases to write your message).\\
    - Output your response in comma separated format of: Inspirational phrase, motivational message generated from phrase.\\
    Below are the 250 phrases to use:\\
    elite athletes often train\\
    pink name brand workout outfit\\
    care to use proper form\\
    $\cdots$
    \end{spacing}
    \vspace{-0.2cm}
    }
    \end{boxA}
    \vspace{-0.5cm}
    \caption{Prompt instructions given to GPT-4. Instructions for crowd-workers (similar to Phrase-GPT) found in \cite{cox2021diverse}.} 
    \label{fig:prompt}
\end{figure}

Large language models (LLMs) can be used to produce creative content \cite{bubeck2023sparks,karinshak2023working,cegin2023chatgpt,lim2023artificial}. For example, Karinshak et al. \cite{karinshak2023working} used GPT-3 to generate messages to persuade people to receive the Covid-19 vaccine.
Furthermore, the quality of LLM output is affected by the prompt used \cite{brown2020language,wu2022ai}.
Specifically, it has been found that LLMs function better with few-shot prompts (i.e., instructions alongside example output) \cite{brown2020language} rather than using zero-shot prompts (with no examples). 
This indicates that crowdsourcing instructions (that often include specific language, examples and atomised tasks) could prove effective as LLM prompts, and some prior work has used instructions from crowd-sourcing tasks to prompt LLMs \cite{wu2023llms,cegin2023chatgpt}.
For example, Cegin et al. \cite{cegin2023chatgpt} found that ChatGPT could paraphrase text in more diverse ways than the crowd, and Wu et al. investigated the ease in adopting multiple crowdsourcing pipelines into LLM prompts \cite{wu2023llms}.
Similarly, recent studies have compared behaviour and task proficiency between humans and LLMs \cite{webson2023language,lampinen2022can}, with some finding LLMs exceed crowd-workers in a number of tasks \cite{gilardi2023chatgpt,hamalainen2023evaluating,tornberg2023chatgpt} (such as data labelling \cite{gilardi2023chatgpt,tornberg2023chatgpt}), while others have found humans more proficient in nuanced creative tasks \cite{stevenson2022putting,jentzsch2023chatgpt}.






Building on this prior work, we investigate the use of a crowdsourcing pipeline to generate a \textit{collectively diverse} corpus of motivational messages to encourage physical activity.
Specifically, we used the \textit{Directed Diversity} \cite{cox2021diverse} pipeline to prompt GPT-4 \cite{bubeck2023sparks} to generate motivational messages (a complex creativity task) by incorporating short example phrases into each message.
We found that GPT-4 messages produced using \cite{cox2021diverse} were more diverse than two baseline GPT-4 prompts. Finally, we discuss insights from messages generated by humans and GPT-4.



\section{Method}


We compared the collective diversity of motivational messages that were generated by human writers and by three different GPT-4 prompts via the ChatGPT web-UI (see Figure \ref{fig:prompt} for GPT-4 prompts).
For the \textbf{Human-Written} condition, we used 250 messages written by crowd-workers (taken from Cox et al. \cite{cox2021diverse}). Here workers were shown one 3 to 5 word phrase to inspire them when writing each message. These phrases were chosen to be semantically diverse in order to create a collectively diverse corpus of messages. We used the same phrases and instructions to prompt GPT-4 to write 250 messages (in the \textbf{Phrase-GPT} condition). For comparison, we also prompted with a \textbf{Simple-GPT} condition: simply asking GPT-4 to write 250 messages; and a \textbf{Diverse-Naïve-GPT} condition: requesting GPT-4 to write 250 messages that are diverse from one another.

\begin{table}[h!]
  \centering
  \hspace*{-0.35cm}
  \begin{tabular}{p{1.55cm}p{3.25cm}p{3.25cm}}
    \toprule
    {\footnotesize Inspirational phrase} & {\footnotesize \textbf{Human-Written} message} & {\footnotesize \textbf{Phrase-GPT} message} \\
    \midrule


{\footnotesize \vspace{-0.27cm} \setword{\textbf{Ex.1}}{Example1} lots of fear and intimidation} &
  {\footnotesize If you can't exercise outside, try listening to music while exercising   inside.  It can help relax and motivate   you at the same time.} &
  {\footnotesize Overcome fear and intimidation in your exercise journey by seeking   support, educating yourself, and taking small steps towards your goals.} \\

{\footnotesize \vspace{-0.27cm} \setword{\textbf{Ex.2}}{Example2} homeopathic doses of exercise} &
  {\footnotesize Even small amounts of exercise can help you become healthy. You don't   need to dedicate 2 hours a day to become healthier!} &
  {\footnotesize Start with small, manageable doses of exercise to build a foundation for a healthier, more active lifestyle.} \\


{\footnotesize \vspace{-0.27cm} \setword{\textbf{Ex.3}}{Example4} military press one rep} &
  {\footnotesize Don't focus on working out for 30 minutes, an hour, etc.} &
  {\footnotesize Just focus on your exercises one rep at a time!	Challenge yourself with exercises like the military press, focusing on form and gradually increasing weight for improved strength.} \\

{\footnotesize \vspace{-0.27cm} \setword{\textbf{Ex.4}}{Example5} rats doing regular aerobic exercise} &
  {\footnotesize Regular running and aerobic exercise is great! Never stop working out.} &
  {\footnotesize Take inspiration from studies, even those on rats, that show the benefits of regular aerobic exercise for overall health and longevity.} \\

{\footnotesize \vspace{-0.27cm} \setword{\textbf{Ex.5}}{Example6} vice president of program development} &
  {\footnotesize Just 30 minutes of exercise a day can change your life!} &
  {\footnotesize Seek guidance from experienced professionals like vice presidents of program development to optimize your fitness routine and approach.} \\
  \bottomrule  
\end{tabular}
  \caption{Example messages written by the crowd and GPT-4 using the same phrases for inspiration.}
  \label{tab:example-messages}
\end{table}

\section{Results}

To calculate the diversity of each set of messages, we calculated the mean pairwise Euclidean distance \cite{kaminskas2016diversity,cox2021diverse} between all messages within each condition (where a higher distance reflects more diversity). From lowest to highest diversity, this gave us: 4.13 for \textbf{Simple-GPT}, 4.29 for \textbf{Naïve-Diverse-GPT}, 5.66 for \textbf{Phrase-GPT}, and 6.90 for \textbf{Human-Written}. This indicates that such a crowdsourcing pipeline could be used to increase the diversity of content generated by LLMs. While Phrase-GPT did not produce a corpus of messages as diverse as those from human-writers, this may be due to differences in message length (with Human-Written averaging 24.0 words, and Phrase-GPT 18.7 words). In addition, Simple-GPT averaged 9.2 and Naïve-Diverse-GPT 9.8 words per message (emphasising the impact of including examples when prompting LLMs).

The work involved in producing messages should also be noted.
The 250 human-written messages took on average 73 seconds each to be written, while GPT-4 took roughly 6 seconds per message.
Additionally, while some human-written messages in \cite{cox2021diverse} were excluded (such as those using poor levels of English or apparent gibberish), the LLM-written messages seemingly suffered from no such issues.
Example Human-Written and Phrase-GPT messages can be found in Table \ref{tab:example-messages} alongside their respective inspirational phrases.


\section{Discussion and Conclusion}

This study has demonstrated the effectiveness of using a crowdsourcing pipeline (\cite{cox2021diverse}) to generate more diverse messages compared to two baseline prompts. However, similar to some previous creativity tasks that require more advanced reasoning abilities \cite{stevenson2022putting,jentzsch2023chatgpt}, human-writers were more successful than GPT-4. Further investigation could alter LLM parameters such as temperature (default 1.0 on ChatGPT \cite{OpenAITemperature}).


Several additional insights are demonstrated by examples in Table \ref{tab:example-messages}. Both human and LLM messages demonstrated the ability to draw metaphors from phrases (\ref{Example2}). GPT-4 may have had difficulty deciding on the relevance of phrases and would generally incorporate phrases, while human writers would act more discerningly (see \ref{Example5} and \ref{Example6}). This emphasises that LLMs follow the form rather than meaning of language \cite{bender2020climbing}, and implies that crowdsourcing pipelines could be atomised further for LLMs (e.g., including an initial step asking the LLM to judge the relevance of a phrase to physical activity).

At times, human-writers would not incorporate phrases (perhaps if they do not have domain knowledge of more esoteric phrases) while the LLM could (\ref{Example4}).
However, risk of AI hallucination (e.g., within healthcare \cite{bubeck2023sparks,lee2023benefits,taylor2022galactica,chen2023utility,jeblick2022chatgpt} such as LLMs misunderstanding medical vocabulary or providing advice that does not follow medical guidelines \cite{chen2023utility,jeblick2022chatgpt}) should be noted, and additional measures would be needed to ensure the veracity of output.


Additionally, (while one could prompt a LLM to incorporate different conversational styles or sociocultural perspectives), attention is needed to ensure that it does not produce harmful cultural stereotypes \cite{durmus2023towards,mirowski2023co,feng2023pretraining}.
Similarly, human writers incorporate personal experiences into messages \cite{coley2013crowdsourced} (\ref{Example1}) that may not necessarily be available to a LLM (if such experiences are poorly represented).

Finally, while we used automated measures to indicate diversity of each experiment condition, further human evaluations for diversity could be conducted \cite{dow2010parallel,cox2021diverse,siangliulue2016ideahound}, in addition to human evaluations of message efficacy (e.g., motivation \cite{de2016crowd,cox2021diverse}).

\begin{acks}
This research is part of the programme DesCartes and is supported by the National Research Foundation, Prime Minister’s Office, Singapore under its Campus for Research Excellence and Technological Enterprise (CREATE) programme.
\end{acks}
\bibliographystyle{ACM-Reference-Format}
\bibliography{sample-base}


\begin{thebibliography}{29}


\ifx \showCODEN    \undefined \def \showCODEN     #1{\unskip}     \fi
\ifx \showDOI      \undefined \def \showDOI       #1{#1}\fi
\ifx \showISBNx    \undefined \def \showISBNx     #1{\unskip}     \fi
\ifx \showISBNxiii \undefined \def \showISBNxiii  #1{\unskip}     \fi
\ifx \showISSN     \undefined \def \showISSN      #1{\unskip}     \fi
\ifx \showLCCN     \undefined \def \showLCCN      #1{\unskip}     \fi
\ifx \shownote     \undefined \def \shownote      #1{#1}          \fi
\ifx \showarticletitle \undefined \def \showarticletitle #1{#1}   \fi
\ifx \showURL      \undefined \def \showURL       {\relax}        \fi
\providecommand\bibfield[2]{#2}
\providecommand\bibinfo[2]{#2}
\providecommand\natexlab[1]{#1}
\providecommand\showeprint[2][]{arXiv:#2}

\bibitem[\protect\citeauthoryear{Bender and Koller}{Bender and Koller}{2020}]%
        {bender2020climbing}
\bibfield{author}{\bibinfo{person}{Emily~M Bender} {and}
  \bibinfo{person}{Alexander Koller}.} \bibinfo{year}{2020}\natexlab{}.
\newblock \showarticletitle{Climbing towards NLU: On meaning, form, and
  understanding in the age of data}. In \bibinfo{booktitle}{\emph{Proceedings
  of the 58th annual meeting of the association for computational
  linguistics}}. \bibinfo{pages}{5185--5198}.
\newblock
\urldef\tempurl%
\url{http://dx.doi.org/10.18653/v1/2020.acl-main.463}
\showURL{%
\tempurl}


\bibitem[\protect\citeauthoryear{Brown, Mann, Ryder, Subbiah, Kaplan, Dhariwal,
  Neelakantan, Shyam, Sastry, Askell, et~al\mbox{.}}{Brown
  et~al\mbox{.}}{2020}]%
        {brown2020language}
\bibfield{author}{\bibinfo{person}{Tom Brown}, \bibinfo{person}{Benjamin Mann},
  \bibinfo{person}{Nick Ryder}, \bibinfo{person}{Melanie Subbiah},
  \bibinfo{person}{Jared~D Kaplan}, \bibinfo{person}{Prafulla Dhariwal},
  \bibinfo{person}{Arvind Neelakantan}, \bibinfo{person}{Pranav Shyam},
  \bibinfo{person}{Girish Sastry}, \bibinfo{person}{Amanda Askell},
  {et~al\mbox{.}}} \bibinfo{year}{2020}\natexlab{}.
\newblock \showarticletitle{Language Models are Few-Shot Learners}.
\newblock \bibinfo{journal}{\emph{Advances in Neural Information Processing
  Systems}}  \bibinfo{volume}{33} (\bibinfo{year}{2020}),
  \bibinfo{pages}{1877--1901}.
\newblock
\newblock
\shownote{\href{https://proceedings.neurips.cc/paper_files/paper/2020/file/1457c0d6bfcb4967418bfb8ac142f64a-Paper.pdf}{NeurIPS
  pdf link}}.


\bibitem[\protect\citeauthoryear{Bubeck, Chandrasekaran, Eldan, Gehrke,
  Horvitz, Kamar, Lee, Lee, Li, Lundberg, et~al\mbox{.}}{Bubeck
  et~al\mbox{.}}{2023}]%
        {bubeck2023sparks}
\bibfield{author}{\bibinfo{person}{S{\'e}bastien Bubeck},
  \bibinfo{person}{Varun Chandrasekaran}, \bibinfo{person}{Ronen Eldan},
  \bibinfo{person}{Johannes Gehrke}, \bibinfo{person}{Eric Horvitz},
  \bibinfo{person}{Ece Kamar}, \bibinfo{person}{Peter Lee},
  \bibinfo{person}{Yin~Tat Lee}, \bibinfo{person}{Yuanzhi Li},
  \bibinfo{person}{Scott Lundberg}, {et~al\mbox{.}}}
  \bibinfo{year}{2023}\natexlab{}.
\newblock \showarticletitle{Sparks of artificial general intelligence: Early
  experiments with gpt-4}.
\newblock \bibinfo{journal}{\emph{arXiv preprint arXiv:2303.12712}}
  (\bibinfo{year}{2023}).
\newblock
\urldef\tempurl%
\url{https://doi.org/10.48550/arXiv.2303.12712}
\showURL{%
\tempurl}


\bibitem[\protect\citeauthoryear{Cegin, Simko, and Brusilovsky}{Cegin
  et~al\mbox{.}}{2023}]%
        {cegin2023chatgpt}
\bibfield{author}{\bibinfo{person}{Jan Cegin}, \bibinfo{person}{Jakub Simko},
  {and} \bibinfo{person}{Peter Brusilovsky}.} \bibinfo{year}{2023}\natexlab{}.
\newblock \showarticletitle{ChatGPT to Replace Crowdsourcing of Paraphrases for
  Intent Classification: Higher Diversity and Comparable Model Robustness}.
\newblock \bibinfo{journal}{\emph{arXiv preprint arXiv:2305.12947}}
  (\bibinfo{year}{2023}).
\newblock
\urldef\tempurl%
\url{https://doi.org/10.48550/arXiv.2305.12947}
\showURL{%
\tempurl}


\bibitem[\protect\citeauthoryear{Chen, Kann, Foote, Aerts, Savova, Mak, and
  Bitterman}{Chen et~al\mbox{.}}{2023}]%
        {chen2023utility}
\bibfield{author}{\bibinfo{person}{Shan Chen}, \bibinfo{person}{Benjamin~H
  Kann}, \bibinfo{person}{Michael~B Foote}, \bibinfo{person}{Hugo~JWL Aerts},
  \bibinfo{person}{Guergana~K Savova}, \bibinfo{person}{Raymond~H Mak}, {and}
  \bibinfo{person}{Danielle~S Bitterman}.} \bibinfo{year}{2023}\natexlab{}.
\newblock \showarticletitle{The utility of ChatGPT for cancer treatment
  information}.
\newblock \bibinfo{journal}{\emph{medRxiv}} (\bibinfo{year}{2023}),
  \bibinfo{pages}{2023--03}.
\newblock
\urldef\tempurl%
\url{https://www.medrxiv.org/content/10.1101/2023.03.16.23287316v1}
\showURL{%
\tempurl}


\bibitem[\protect\citeauthoryear{Coley, Sadasivam, Williams, Volkman,
  Schoenberger, Kohler, Sobko, Ray, Allison, Ford, et~al\mbox{.}}{Coley
  et~al\mbox{.}}{2013}]%
        {coley2013crowdsourced}
\bibfield{author}{\bibinfo{person}{Heather~L Coley}, \bibinfo{person}{Rajani~S
  Sadasivam}, \bibinfo{person}{Jessica~H Williams}, \bibinfo{person}{Julie~E
  Volkman}, \bibinfo{person}{Yu-Mei Schoenberger}, \bibinfo{person}{Connie~L
  Kohler}, \bibinfo{person}{Heather Sobko}, \bibinfo{person}{Midge~N Ray},
  \bibinfo{person}{Jeroan~J Allison}, \bibinfo{person}{Daniel~E Ford},
  {et~al\mbox{.}}} \bibinfo{year}{2013}\natexlab{}.
\newblock \showarticletitle{Crowdsourced Peer- Versus Expert-Written
  Smoking-Cessation Messages}.
\newblock \bibinfo{journal}{\emph{American journal of preventive medicine}}
  \bibinfo{volume}{45}, \bibinfo{number}{5} (\bibinfo{year}{2013}),
  \bibinfo{pages}{543--550}.
\newblock
\urldef\tempurl%
\url{https://doi.org/10.1016/j.amepre.2013.07.004}
\showURL{%
\tempurl}


\bibitem[\protect\citeauthoryear{Cox, Wang, Abdul, von~der Weth, and
  Y.~Lim}{Cox et~al\mbox{.}}{2021}]%
        {cox2021diverse}
\bibfield{author}{\bibinfo{person}{Samuel~Rhys Cox}, \bibinfo{person}{Yunlong
  Wang}, \bibinfo{person}{Ashraf Abdul}, \bibinfo{person}{Christian von~der
  Weth}, {and} \bibinfo{person}{Brian Y.~Lim}.}
  \bibinfo{year}{2021}\natexlab{}.
\newblock \showarticletitle{Directed Diversity: Leveraging Language Embedding
  Distances for Collective Creativity in Crowd Ideation}. In
  \bibinfo{booktitle}{\emph{Proceedings of the 2021 CHI Conference on Human
  Factors in Computing Systems}}.
\newblock
\urldef\tempurl%
\url{https://doi.org/10.1145/3411764.3445782}
\showURL{%
\tempurl}


\bibitem[\protect\citeauthoryear{de~Vries, Truong, Kwint, Drossaert, and
  Evers}{de~Vries et~al\mbox{.}}{2016}]%
        {de2016crowd}
\bibfield{author}{\bibinfo{person}{Roelof~AJ de Vries},
  \bibinfo{person}{Khiet~P Truong}, \bibinfo{person}{Sigrid Kwint},
  \bibinfo{person}{Constance~HC Drossaert}, {and} \bibinfo{person}{Vanessa
  Evers}.} \bibinfo{year}{2016}\natexlab{}.
\newblock \showarticletitle{Crowd-Designed Motivation: Motivational Messages
  for Exercise Adherence Based on Behavior Change Theory}. In
  \bibinfo{booktitle}{\emph{Proceedings of the 2016 CHI Conference on Human
  Factors in Computing Systems}}. ACM, \bibinfo{pages}{297--308}.
\newblock
\urldef\tempurl%
\url{https://doi.org/10.1145/2858036.2858229}
\showURL{%
\tempurl}


\bibitem[\protect\citeauthoryear{Dow, Glassco, Kass, Schwarz, Schwartz, and
  Klemmer}{Dow et~al\mbox{.}}{2010}]%
        {dow2010parallel}
\bibfield{author}{\bibinfo{person}{Steven~P Dow}, \bibinfo{person}{Alana
  Glassco}, \bibinfo{person}{Jonathan Kass}, \bibinfo{person}{Melissa Schwarz},
  \bibinfo{person}{Daniel~L Schwartz}, {and} \bibinfo{person}{Scott~R
  Klemmer}.} \bibinfo{year}{2010}\natexlab{}.
\newblock \showarticletitle{Parallel prototyping leads to better design
  results, more divergence, and increased self-efficacy}.
\newblock \bibinfo{journal}{\emph{ACM Transactions on Computer-Human
  Interaction (TOCHI)}} \bibinfo{volume}{17}, \bibinfo{number}{4}
  (\bibinfo{year}{2010}), \bibinfo{pages}{18}.
\newblock
\urldef\tempurl%
\url{https://doi.org/10.1145/1879831.1879836}
\showURL{%
\tempurl}


\bibitem[\protect\citeauthoryear{Durmus, Nyugen, Liao, Schiefer, Askell,
  Bakhtin, Chen, Hatfield-Dodds, Hernandez, Joseph, et~al\mbox{.}}{Durmus
  et~al\mbox{.}}{2023}]%
        {durmus2023towards}
\bibfield{author}{\bibinfo{person}{Esin Durmus}, \bibinfo{person}{Karina
  Nyugen}, \bibinfo{person}{Thomas~I Liao}, \bibinfo{person}{Nicholas
  Schiefer}, \bibinfo{person}{Amanda Askell}, \bibinfo{person}{Anton Bakhtin},
  \bibinfo{person}{Carol Chen}, \bibinfo{person}{Zac Hatfield-Dodds},
  \bibinfo{person}{Danny Hernandez}, \bibinfo{person}{Nicholas Joseph},
  {et~al\mbox{.}}} \bibinfo{year}{2023}\natexlab{}.
\newblock \showarticletitle{Towards Measuring the Representation of Subjective
  Global Opinions in Language Models}.
\newblock \bibinfo{journal}{\emph{arXiv preprint arXiv:2306.16388}}
  (\bibinfo{year}{2023}).
\newblock
\urldef\tempurl%
\url{https://doi.org/10.48550/arXiv.2306.16388}
\showURL{%
\tempurl}


\bibitem[\protect\citeauthoryear{Feng, Park, Liu, and Tsvetkov}{Feng
  et~al\mbox{.}}{2023}]%
        {feng2023pretraining}
\bibfield{author}{\bibinfo{person}{Shangbin Feng}, \bibinfo{person}{Chan~Young
  Park}, \bibinfo{person}{Yuhan Liu}, {and} \bibinfo{person}{Yulia Tsvetkov}.}
  \bibinfo{year}{2023}\natexlab{}.
\newblock \showarticletitle{From Pretraining Data to Language Models to
  Downstream Tasks: Tracking the Trails of Political Biases Leading to Unfair
  {NLP} Models}. In \bibinfo{booktitle}{\emph{Proceedings of the 61st Annual
  Meeting of the Association for Computational Linguistics (Volume 1: Long
  Papers)}}. \bibinfo{publisher}{Association for Computational Linguistics},
  \bibinfo{address}{Toronto, Canada}, \bibinfo{pages}{11737--11762}.
\newblock
\urldef\tempurl%
\url{https://doi.org/10.18653/v1/2023.acl-long.656}
\showDOI{\tempurl}


\bibitem[\protect\citeauthoryear{Gilardi, Alizadeh, and Kubli}{Gilardi
  et~al\mbox{.}}{2023}]%
        {gilardi2023chatgpt}
\bibfield{author}{\bibinfo{person}{Fabrizio Gilardi}, \bibinfo{person}{Meysam
  Alizadeh}, {and} \bibinfo{person}{Maël Kubli}.}
  \bibinfo{year}{2023}\natexlab{}.
\newblock \showarticletitle{ChatGPT outperforms crowd workers for
  text-annotation tasks}.
\newblock \bibinfo{journal}{\emph{Proceedings of the National Academy of
  Sciences}} \bibinfo{volume}{120}, \bibinfo{number}{30}
  (\bibinfo{year}{2023}), \bibinfo{pages}{e2305016120}.
\newblock
\urldef\tempurl%
\url{https://doi.org/10.1073/pnas.2305016120}
\showDOI{\tempurl}


\bibitem[\protect\citeauthoryear{H{\"a}m{\"a}l{\"a}inen, Tavast, and
  Kunnari}{H{\"a}m{\"a}l{\"a}inen et~al\mbox{.}}{2023}]%
        {hamalainen2023evaluating}
\bibfield{author}{\bibinfo{person}{Perttu H{\"a}m{\"a}l{\"a}inen},
  \bibinfo{person}{Mikke Tavast}, {and} \bibinfo{person}{Anton Kunnari}.}
  \bibinfo{year}{2023}\natexlab{}.
\newblock \showarticletitle{Evaluating Large Language Models in Generating
  Synthetic HCI Research Data: a Case Study}. In
  \bibinfo{booktitle}{\emph{Proceedings of the 2023 CHI Conference on Human
  Factors in Computing Systems}}. \bibinfo{pages}{1--19}.
\newblock
\urldef\tempurl%
\url{https://doi.org/10.1145/3544548.3580688}
\showURL{%
\tempurl}


\bibitem[\protect\citeauthoryear{Jeblick, Schachtner, Dexl, Mittermeier,
  St{\"u}ber, Topalis, Weber, Wesp, Sabel, Ricke, et~al\mbox{.}}{Jeblick
  et~al\mbox{.}}{2022}]%
        {jeblick2022chatgpt}
\bibfield{author}{\bibinfo{person}{Katharina Jeblick},
  \bibinfo{person}{Balthasar Schachtner}, \bibinfo{person}{Jakob Dexl},
  \bibinfo{person}{Andreas Mittermeier}, \bibinfo{person}{Anna~Theresa
  St{\"u}ber}, \bibinfo{person}{Johanna Topalis}, \bibinfo{person}{Tobias
  Weber}, \bibinfo{person}{Philipp Wesp}, \bibinfo{person}{Bastian Sabel},
  \bibinfo{person}{Jens Ricke}, {et~al\mbox{.}}}
  \bibinfo{year}{2022}\natexlab{}.
\newblock \showarticletitle{ChatGPT Makes Medicine Easy to Swallow: An
  Exploratory Case Study on Simplified Radiology Reports}.
\newblock \bibinfo{journal}{\emph{arXiv preprint arXiv:2212.14882}}
  (\bibinfo{year}{2022}).
\newblock
\urldef\tempurl%
\url{https://doi.org/10.48550/arXiv.2212.14882}
\showURL{%
\tempurl}


\bibitem[\protect\citeauthoryear{Jentzsch and Kersting}{Jentzsch and
  Kersting}{2023}]%
        {jentzsch2023chatgpt}
\bibfield{author}{\bibinfo{person}{Sophie Jentzsch} {and}
  \bibinfo{person}{Kristian Kersting}.} \bibinfo{year}{2023}\natexlab{}.
\newblock \showarticletitle{ChatGPT is fun, but it is not funny! Humor is still
  challenging Large Language Models}.
\newblock \bibinfo{journal}{\emph{arXiv preprint arXiv:2306.04563}}
  (\bibinfo{year}{2023}).
\newblock
\urldef\tempurl%
\url{https://doi.org/10.48550/arXiv.2306.04563}
\showURL{%
\tempurl}


\bibitem[\protect\citeauthoryear{Kaminskas and Bridge}{Kaminskas and
  Bridge}{2016}]%
        {kaminskas2016diversity}
\bibfield{author}{\bibinfo{person}{Marius Kaminskas} {and}
  \bibinfo{person}{Derek Bridge}.} \bibinfo{year}{2016}\natexlab{}.
\newblock \showarticletitle{Diversity, Serendipity, Novelty, and Coverage: A
  Survey and Empirical Analysis of Beyond-Accuracy Objectives in Recommender
  Systems}.
\newblock \bibinfo{journal}{\emph{ACM Transactions on Interactive Intelligent
  Systems (TiiS)}} \bibinfo{volume}{7}, \bibinfo{number}{1}
  (\bibinfo{year}{2016}), \bibinfo{pages}{1--42}.
\newblock
\urldef\tempurl%
\url{https://doi.org/10.1145/2926720}
\showURL{%
\tempurl}


\bibitem[\protect\citeauthoryear{Karinshak, Liu, Park, and Hancock}{Karinshak
  et~al\mbox{.}}{2023}]%
        {karinshak2023working}
\bibfield{author}{\bibinfo{person}{Elise Karinshak}, \bibinfo{person}{Sunny~Xun
  Liu}, \bibinfo{person}{Joon~Sung Park}, {and} \bibinfo{person}{Jeffrey~T
  Hancock}.} \bibinfo{year}{2023}\natexlab{}.
\newblock \showarticletitle{Working With AI to Persuade: Examining a Large
  Language Model's Ability to Generate Pro-Vaccination Messages}.
\newblock \bibinfo{journal}{\emph{Proceedings of the ACM on Human-Computer
  Interaction}} \bibinfo{volume}{7}, \bibinfo{number}{CSCW1}
  (\bibinfo{year}{2023}), \bibinfo{pages}{1--29}.
\newblock
\urldef\tempurl%
\url{https://doi.org/10.1145/3579592}
\showURL{%
\tempurl}


\bibitem[\protect\citeauthoryear{Lampinen}{Lampinen}{2022}]%
        {lampinen2022can}
\bibfield{author}{\bibinfo{person}{Andrew~Kyle Lampinen}.}
  \bibinfo{year}{2022}\natexlab{}.
\newblock \showarticletitle{Can language models handle recursively nested
  grammatical structures? A case study on comparing models and humans}.
\newblock \bibinfo{journal}{\emph{arXiv preprint arXiv:2210.15303}}
  (\bibinfo{year}{2022}).
\newblock
\urldef\tempurl%
\url{https://doi.org/10.48550/arXiv.2210.15303}
\showURL{%
\tempurl}


\bibitem[\protect\citeauthoryear{Lee, Bubeck, and Petro}{Lee
  et~al\mbox{.}}{2023}]%
        {lee2023benefits}
\bibfield{author}{\bibinfo{person}{Peter Lee}, \bibinfo{person}{Sebastien
  Bubeck}, {and} \bibinfo{person}{Joseph Petro}.}
  \bibinfo{year}{2023}\natexlab{}.
\newblock \showarticletitle{Benefits, Limits, and Risks of GPT-4 as an AI
  Chatbot for Medicine}.
\newblock \bibinfo{journal}{\emph{New England Journal of Medicine}}
  \bibinfo{volume}{388}, \bibinfo{number}{13} (\bibinfo{year}{2023}),
  \bibinfo{pages}{1233--1239}.
\newblock
\urldef\tempurl%
\url{https://doi.org/10.1056/NEJMsr2214184}
\showURL{%
\tempurl}


\bibitem[\protect\citeauthoryear{Lim and Schm{\"a}lzle}{Lim and
  Schm{\"a}lzle}{2023}]%
        {lim2023artificial}
\bibfield{author}{\bibinfo{person}{Sue Lim} {and} \bibinfo{person}{Ralf
  Schm{\"a}lzle}.} \bibinfo{year}{2023}\natexlab{}.
\newblock \showarticletitle{Artificial intelligence for health message
  generation: an empirical study using a large language model (LLM) and prompt
  engineering}.
\newblock \bibinfo{journal}{\emph{Frontiers in Communication}}
  \bibinfo{volume}{8} (\bibinfo{year}{2023}), \bibinfo{pages}{1129082}.
\newblock
\urldef\tempurl%
\url{https://doi.org/10.3389/fcomm.2023.1129082}
\showURL{%
\tempurl}


\bibitem[\protect\citeauthoryear{Mirowski, Mathewson, Pittman, and
  Evans}{Mirowski et~al\mbox{.}}{2023}]%
        {mirowski2023co}
\bibfield{author}{\bibinfo{person}{Piotr Mirowski}, \bibinfo{person}{Kory~W
  Mathewson}, \bibinfo{person}{Jaylen Pittman}, {and} \bibinfo{person}{Richard
  Evans}.} \bibinfo{year}{2023}\natexlab{}.
\newblock \showarticletitle{Co-Writing Screenplays and Theatre Scripts with
  Language Models: Evaluation by Industry Professionals}. In
  \bibinfo{booktitle}{\emph{Proceedings of the 2023 CHI Conference on Human
  Factors in Computing Systems}}. \bibinfo{pages}{1--34}.
\newblock
\urldef\tempurl%
\url{https://doi.org/10.1145/3544548.3581225}
\showURL{%
\tempurl}


\bibitem[\protect\citeauthoryear{OpenAI}{OpenAI}{2023}]%
        {OpenAITemperature}
\bibfield{author}{\bibinfo{person}{OpenAI}.} \bibinfo{year}{2023}\natexlab{}.
\newblock \showarticletitle{{API Reference - OpenAI API}}.
\newblock  (\bibinfo{year}{2023}).
\newblock
\urldef\tempurl%
\url{https://platform.openai.com/docs/api-reference/chat/create#chat/create-temperature}
\showURL{%
\tempurl}


\bibitem[\protect\citeauthoryear{Siangliulue, Chan, Dow, and Gajos}{Siangliulue
  et~al\mbox{.}}{2016}]%
        {siangliulue2016ideahound}
\bibfield{author}{\bibinfo{person}{Pao Siangliulue}, \bibinfo{person}{Joel
  Chan}, \bibinfo{person}{Steven~P Dow}, {and} \bibinfo{person}{Krzysztof~Z
  Gajos}.} \bibinfo{year}{2016}\natexlab{}.
\newblock \showarticletitle{IdeaHound: Improving Large-scale Collaborative
  Ideation with Crowd-Powered Real-time Semantic Modeling}. In
  \bibinfo{booktitle}{\emph{Proceedings of the 29th Annual Symposium on User
  Interface Software and Technology}}. \bibinfo{pages}{609--624}.
\newblock
\urldef\tempurl%
\url{https://doi.org/10.1145/2984511.2984578}
\showURL{%
\tempurl}


\bibitem[\protect\citeauthoryear{Stevenson, Smal, Baas, Grasman, and van~der
  Maas}{Stevenson et~al\mbox{.}}{2022}]%
        {stevenson2022putting}
\bibfield{author}{\bibinfo{person}{Claire Stevenson}, \bibinfo{person}{Iris
  Smal}, \bibinfo{person}{Matthijs Baas}, \bibinfo{person}{Raoul Grasman},
  {and} \bibinfo{person}{Han van~der Maas}.} \bibinfo{year}{2022}\natexlab{}.
\newblock \showarticletitle{Putting GPT-3's Creativity to the (Alternative
  Uses) Test}.
\newblock \bibinfo{journal}{\emph{arXiv preprint arXiv:2206.08932}}
  (\bibinfo{year}{2022}).
\newblock
\urldef\tempurl%
\url{https://doi.org/10.48550/arXiv.2206.08932}
\showURL{%
\tempurl}


\bibitem[\protect\citeauthoryear{Taylor, Kardas, Cucurull, Scialom, Hartshorn,
  Saravia, Poulton, Kerkez, and Stojnic}{Taylor et~al\mbox{.}}{2022}]%
        {taylor2022galactica}
\bibfield{author}{\bibinfo{person}{Ross Taylor}, \bibinfo{person}{Marcin
  Kardas}, \bibinfo{person}{Guillem Cucurull}, \bibinfo{person}{Thomas
  Scialom}, \bibinfo{person}{Anthony Hartshorn}, \bibinfo{person}{Elvis
  Saravia}, \bibinfo{person}{Andrew Poulton}, \bibinfo{person}{Viktor Kerkez},
  {and} \bibinfo{person}{Robert Stojnic}.} \bibinfo{year}{2022}\natexlab{}.
\newblock \showarticletitle{Galactica: A Large Language Model for Science}.
\newblock \bibinfo{journal}{\emph{arXiv preprint arXiv:2211.09085}}
  (\bibinfo{year}{2022}).
\newblock
\urldef\tempurl%
\url{https://doi.org/10.48550/arXiv.2211.09085}
\showURL{%
\tempurl}


\bibitem[\protect\citeauthoryear{T{\"o}rnberg}{T{\"o}rnberg}{2023}]%
        {tornberg2023chatgpt}
\bibfield{author}{\bibinfo{person}{Petter T{\"o}rnberg}.}
  \bibinfo{year}{2023}\natexlab{}.
\newblock \showarticletitle{ChatGPT-4 Outperforms Experts and Crowd Workers in
  Annotating Political Twitter Messages with Zero-Shot Learning}.
\newblock \bibinfo{journal}{\emph{arXiv preprint arXiv:2304.06588}}
  (\bibinfo{year}{2023}).
\newblock
\urldef\tempurl%
\url{https://doi.org/10.48550/arXiv.2304.06588}
\showURL{%
\tempurl}


\bibitem[\protect\citeauthoryear{Webson, Loo, Yu, and Pavlick}{Webson
  et~al\mbox{.}}{2023}]%
        {webson2023language}
\bibfield{author}{\bibinfo{person}{Albert Webson},
  \bibinfo{person}{Alyssa~Marie Loo}, \bibinfo{person}{Qinan Yu}, {and}
  \bibinfo{person}{Ellie Pavlick}.} \bibinfo{year}{2023}\natexlab{}.
\newblock \showarticletitle{Are Language Models Worse than Humans at Following
  Prompts? It's Complicated}.
\newblock \bibinfo{journal}{\emph{arXiv preprint arXiv:2301.07085}}
  (\bibinfo{year}{2023}).
\newblock
\urldef\tempurl%
\url{https://doi.org/10.48550/arXiv.2301.07085}
\showURL{%
\tempurl}


\bibitem[\protect\citeauthoryear{Wu, Terry, and Cai}{Wu et~al\mbox{.}}{2022}]%
        {wu2022ai}
\bibfield{author}{\bibinfo{person}{Tongshuang Wu}, \bibinfo{person}{Michael
  Terry}, {and} \bibinfo{person}{Carrie~Jun Cai}.}
  \bibinfo{year}{2022}\natexlab{}.
\newblock \showarticletitle{AI Chains: Transparent and Controllable Human-AI
  Interaction by Chaining Large Language Model Prompts}. In
  \bibinfo{booktitle}{\emph{Proceedings of the 2022 CHI conference on human
  factors in computing systems}}. \bibinfo{pages}{1--22}.
\newblock
\urldef\tempurl%
\url{https://doi.org/10.1145/3491102.3517582}
\showURL{%
\tempurl}


\bibitem[\protect\citeauthoryear{Wu, Zhu, Albayrak, Axon, Bertsch, Deng, Ding,
  Guo, Gururaja, Kuo, et~al\mbox{.}}{Wu et~al\mbox{.}}{2023}]%
        {wu2023llms}
\bibfield{author}{\bibinfo{person}{Tongshuang Wu}, \bibinfo{person}{Haiyi Zhu},
  \bibinfo{person}{Maya Albayrak}, \bibinfo{person}{Alexis Axon},
  \bibinfo{person}{Amanda Bertsch}, \bibinfo{person}{Wenxing Deng},
  \bibinfo{person}{Ziqi Ding}, \bibinfo{person}{Bill Guo},
  \bibinfo{person}{Sireesh Gururaja}, \bibinfo{person}{Tzu-Sheng Kuo},
  {et~al\mbox{.}}} \bibinfo{year}{2023}\natexlab{}.
\newblock \showarticletitle{LLMs as Workers in Human-Computational Algorithms?
  Replicating Crowdsourcing Pipelines with LLMs}.
\newblock \bibinfo{journal}{\emph{arXiv preprint arXiv:2307.10168}}
  (\bibinfo{year}{2023}).
\newblock
\urldef\tempurl%
\url{https://doi.org/10.48550/arXiv.2307.10168}
\showURL{%
\tempurl}


\end{thebibliography}



\end{document}